\documentclass[letterpaper, 10 pt, conference]{ieeeconf}  

\IEEEoverridecommandlockouts                              

\usepackage{amsmath} 
\usepackage{amssymb}  
\usepackage{booktabs}
\usepackage{graphicx}
\usepackage{subfig}
\usepackage{cite}
\usepackage{xurl}
\usepackage[hidelinks]{hyperref}
\usepackage{cleveref}
\crefname{figure}{Fig.}{Figs.}
\Crefname{figure}{Fig.}{Figs.}

\crefname{table}{Table}{Tables}
\Crefname{table}{Table}{Tables}

\title{\LARGE \bf
PneuGen: Task-Driven Agentic Co-Design of Soft Pneumatic Robots
}

\author{Shaohong Zhong$^{1}$, Marco Pontin$^{1}$, Perla Maiolino$^{1}$%
\thanks{$^{1}$Oxford Robotics Institute, Department of Engineering Science, University of Oxford, OX1 3PJ Oxford, U.K. 
        {\tt\small shaohong@robots.ox.ac.uk}}}%

\begin{document}
\bstctlcite{IEEEcontrol}

\maketitle
\thispagestyle{empty}
\pagestyle{empty}


\begin{abstract}
Task-driven design of soft robots requires the consideration of factors such as geometry, morphology, material properties, and actuation, resulting in a large and complex design space that is challenging to explore. Computational optimisation methods leverage signals from simulation to iteratively improve designs to achieve task objectives. However, these methods typically demand high computational load for both optimisation and simulation. We present PneuGen, an agentic co-design framework for soft robots that leverages Large Language Model (LLM) agents to search over the soft robot design space efficiently. A designer agent proposes design programs from a parametric component library, and a separate critic agent analyses simulation diagnostics and recommends targeted revisions. Fitted module response models provide reusable physical priors relating geometry and pressure to deformation. Design programs are automatically validated and converted into finite-element simulations, enabling iterative evaluation. Across grasping tasks on primitive shapes and challenging objects, as well as shape-matching tasks that require multi-directional planar and 3D deformation, the proposed framework significantly outperforms prior computational optimisation baselines. Ablation experiments also highlight the utility of the critic agent and the module response model in facilitating efficient design. We additionally demonstrate the agent-generated soft robot designs in real-world shape-matching and grasping tasks.

\end{abstract}

\section{Introduction}

Soft robots offer advantages of mechanical compliance, adaptability, and flexibility to achieve a wide range of behaviours, allowing them to accommodate uncertainty in tasks such as manipulation~\cite{shintake2018soft,zhong2026pneutac}. However, their behaviour depends strongly on geometry, material properties, and the arrangement of compliant elements. These make the design of soft robots particularly challenging because of the large and complex design space, where morphology and actuation need to be selected jointly~\cite{pinskier2022bioinspiration}. This coupling motivates task-driven design methods that select geometry and actuation together. In this aspect, heuristic methods rely heavily on the designer’s experience, which limits their scalability~\cite{yao2026generalized}. 

Computational methods have shown great potential in soft robot design~\cite{pinskier2022bioinspiration,yao2023design,yao2024spada,yao2026generalized}. Prior approaches have explored both the optimisation of existing actuator families and the generation of new morphologies~\cite{yao2023design,yao2024spada,yao2026generalized,enkera2026simto}. Parametric methods optimise geometric and actuation variables using physical models or learned surrogates, enabling task-specific design of pneumatic actuators and tendon-driven soft hands~\cite{yao2026generalized}. Shape and topology optimisation extend the search to material distributions and structural features, while graph-based evolutionary methods explore alternative arrangements and connectivity\cite{farinha2026graphbased}. Differentiable simulation and learned physical models further support the joint optimisation of morphology and actuation or grasp configuration~\cite{ma2021diffaqua,zhong2025endtoend}. These advances demonstrate the potential of computational search across increasingly expressive design spaces. However, these methods typically have substantial computational requirements, not only in the design optimisation process, but also through the repeated simulation and evaluation that are required, motivating methods that extract more useful guidance from each evaluation. Furthermore, many of these computational approaches rely on manually crafted loss functions for optimisation, which both limit their generalisability and sometimes do not offer sufficient appropriate signals~\cite{ma2021diffaqua,zhong2025endtoend}. 

\begin{figure} [t]
	\centering
    \includegraphics[width=\columnwidth]{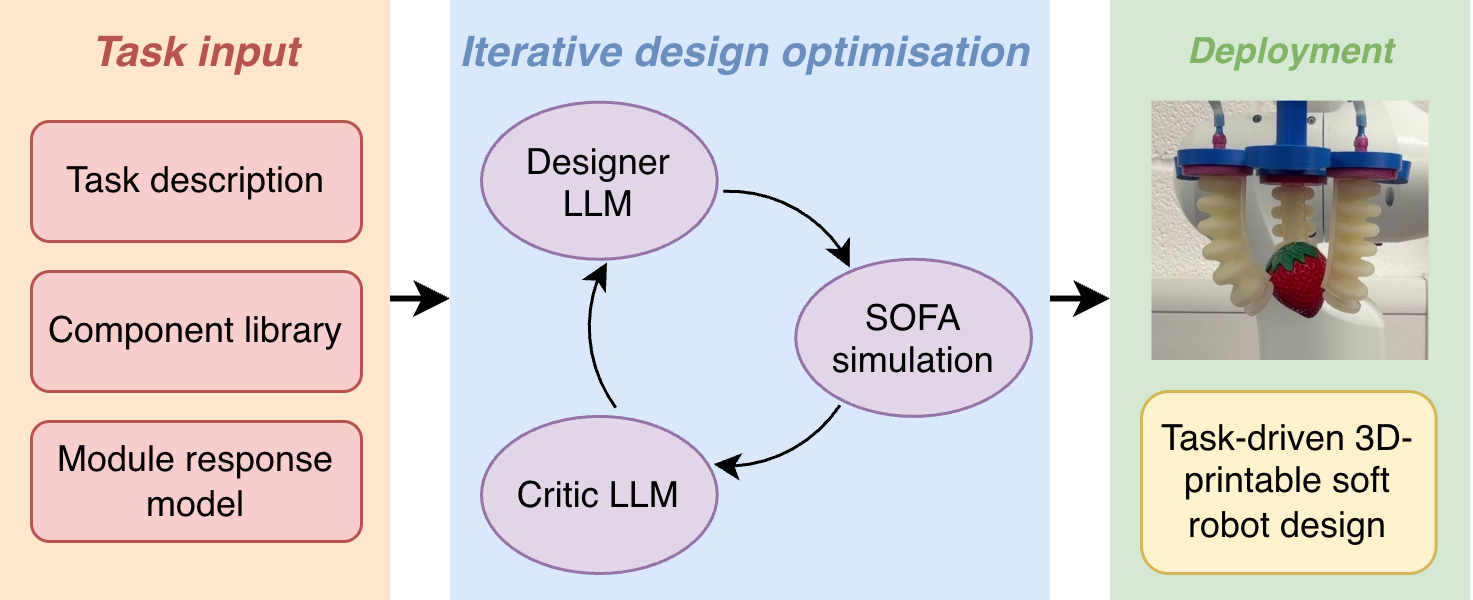}
	\caption{Overview of the PneuGen agentic co-design framework. Given the task inputs, the framework leverages a designer LLM agent and a critic agent to iteratively improve the soft robot design proposals. Final proposals can be directly 3D printed for deployment.}
	\label{fig:overview}
\end{figure} 

Large language models (LLMs) offer a means of incorporating broad prior knowledge into computational design~\cite{stella2023how,qiao2025lang2morph,chen2026bridging}. Through large-scale pretraining, these models are able to interpret task requirements, module behaviour descriptions, and feedback diagnostics. These capabilities help connect a task description to appropriate design choices. Their ability to generate structured descriptions and code also allows these choices to be expressed as design programs and revised in a principled manner within an iterative search. Beyond proposing designs, multimodal LLMs can analyse simulation observations and failure diagnostics to formulate explanations for unsuccessful designs and provide much richer guidance than task scores alone, as well as propose targeted revisions. By employing different instances of LLMs as agents with different roles, we are able to construct a designer-critic paradigm in which appropriate designs can be devised efficiently without extensive computational evaluations and expensive simulation. 

In this paper, we present PneuGen, a task-driven agentic co-design framework for soft robots. An overview of the framework is shown in \cref{fig:overview}. A designer LLM agent constructs designs from a library of parametric components and jointly selects their spatial arrangement, geometric parameters, and actuation. A separate critic agent interprets simulation diagnostics and proposes revisions for subsequent design iterations. Component-level pressure-to-displacement maps, represented by fitted module response models, provide a physics prior on how geometry and pressure affect deformation. The resulting design programs are automatically converted into CAD models, meshes, and finite-element simulations. The framework also supports selection from different families of soft robot components to satisfy task requirements. Our contributions are:
\begin{enumerate}
    \item An agentic co-design loop that combines soft robot geometry and actuation design with simulation-based evaluation and critic-guided refinement.
    \item A library of soft robot components with fitted module response models, geometric and fabrication constraints, and automatic CAD and simulation generation to support task-driven design.
    \item An evaluation on simulated grasping and shape-matching tasks against random search, genetic search, and CMA-ES with real-world demonstrations, and an investigation of the effects of module response models and critic guidance. 
\end{enumerate}

\section{Literature Review}

Our work connects computational optimisation of soft robot morphology and actuation with LLM-based design generation and refinement. In this section, we review these areas with particular attention to the computational cost and the generalisability of the approaches.

\subsection{Computational design optimisation for soft robots}

Prior works have employed computational optimisation methods for the design of soft robots, where the optimisation encompasses parameters, morphology and control. Parametric approaches search within an explicitly defined actuator family. For example, SPADA combines surrogate models derived from finite-element models (FEMs), shape segmentation, and genetic optimisation to generate actuator geometries and export fabrication-ready CAD models~\cite{yao2024spada}. Subsequent work embeds reduced-order FEM surrogates in a rigid-body simulation environment, enabling optimisation of gripper geometry and pressure commands under different conditions~\cite{yao2026generalized}. However, these works typically focus on a single actuator family with pre-defined structures~\cite{yao2023design,yao2024spada}. Additional works attempt to modify structural features and material distributions. For example, SimTO estimates loading conditions from simulated gripper–object interactions and uses them to optimise finger topology, and Farinha et al. combine graph-encoded soft structures with multi-objective quality-diversity optimisation across grasping tasks~\cite{enkera2026simto,farinha2026graphbased}. Additionally, DiffAqua uses differentiable simulation to co-design soft-swimmer geometry and control. These methods broaden the design space, but their search strategies depend on suitable numerical encodings and, for gradient-based optimisation, differentiable representations~\cite{ma2021diffaqua}. They also demand heavy computation, and the optimisation objectives are typically manually crafted and tuned for each task. 

The cost of evaluating candidate designs has also motivated the development of efficient evaluation methods. For example, neural physics models support optimisation of stiffness distributions and grasp poses, while learned reward models reduce the simulations required to search for soft robot designs~\cite{yi2025codesign,bai2025learning}. Genetic search can also reuse policies across related morphologies~\cite{mannam2024design}. However, these learned methods depend on data covering the relevant designs and behaviours, adding a computational burden and limiting generalisability. 

\subsection{Agentic robot design using LLMs}

LLMs have been used to directly generate design proposals and interpret evaluation feedback. For example, RoboMorph generates grammar-constrained modular robots within an evolutionary loop~\cite{qiu2026robomorph}. LASeR incorporates reflection on search trajectories for voxel-based soft robot design, while Auto-Robotist maintains reusable, evidence-supported design rules~\cite{song2025laser,wang2026when}. However, these methods typically focus on simple or planar designs without complex geometries. 

Additionally, LLMs have also been connected to parametric CAD representations. For example, Lang2Morph connects task descriptions to parametric hand geometry, using semantic alignment and dimensional compatibility for refinement~\cite{qiao2025lang2morph}. However, it only uses morphology evaluation, which does not establish performance under physical contact or actual deployment conditions. AID-SR provides this physical grounding through structured simulation feedback for tendon-driven continuum robots~\cite{chen2026bridging}. However, its functionality evaluation requires a reinforcement-learning controller for each candidate, thus coupling the design score to controller training outcomes. The designs also mainly focus on simple structures. 

Compared to the prior methods, the PneuGen framework leverages agentic design for complex soft robots with critic-guided refinement, supporting multiple families of actuators, and also achieving generalisability and efficiency with minimal human intervention. 
\section{Methodology}

\begin{figure*} [ht]
	\centering
	\includegraphics[width=\textwidth]{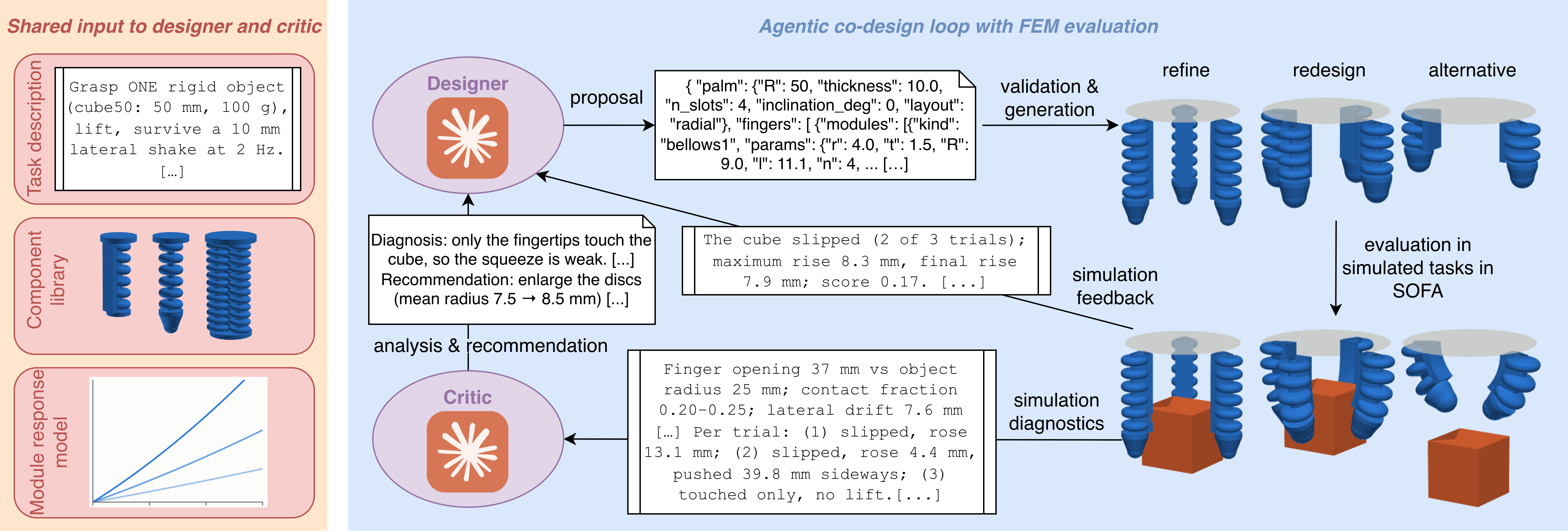}
	\caption{Detailed overview of the PneuGen framework with example excerpts. Left: inputs to the agentic design pipeline. Right: agentic co-design loop. The designer proposes soft robot designs, and each proposal is evaluated in SOFA simulation. The results are fed back to the designer, and to a critic agent that provides an additional in-depth analysis.}
	\label{fig:method}
\end{figure*}

To address the challenges with soft robot design, we propose an agentic task-driven co-design framework that jointly selects soft robot components, their spatial arrangement, geometric parameters, and actuation commands. Given a desired task, a designer LLM agent generates robot design proposals, while a separate critic agent interprets simulation diagnostics and recommends revisions. The designs are evaluated in finite-element simulation for downstream task performance. In this section, we discuss the representation of the design with the module response models, and the agentic designer-critic architecture employed by the PneuGen framework. A detailed overview of the framework is shown in \cref{fig:method}.

\subsection{Design representation with module response model}

In this paper, a candidate soft robot design proposal is represented as
$d=(\mathbf m,\mathcal L,\boldsymbol\theta,\mathbf u)$
where $\mathbf m$ specifies component types, $\mathcal L$ their spatial arrangement, $\boldsymbol\theta$ their geometric parameters, and $\mathbf u$ the actuation commands. Given a task description, the designer then needs to jointly select these values. An additional programmatic validator first checks geometric compatibility, manufacturing constraints, component intersections, and simulation resource limits in the proposed design. Valid design programs are automatically converted into CAD geometry, and tetrahedralised to generate a finite-element mesh. The mesh and actuation specifications are then used to assemble the simulation scene, enabling candidate designs to be evaluated without manual geometry reconstruction or meshing. Validation failures return explicit rejection messages for subsequent refinement.

Additionally, to reduce repeated exploration of module behaviour, we first characterise library components through actuation sweeps and fit compact models relating geometry and pressure to deformation. Let $\mathbf q$ denote a component’s geometric parameters, $\mathbf p$ its actuation inputs, and $\mathbf y$ a vector describing its deformation. We approximate the module response using $\widehat{\mathbf y} = f_{\boldsymbol\beta}(\mathbf q,\mathbf p)$ where $\boldsymbol\beta$ contains coefficients fitted to simulation data collected through actuation sweeps. The agents receive these fitted relationships as reusable physical priors for proposing and revising design changes together with reference examples.

\subsection{Agentic designer–critic architecture}

LLM agents support separate role definitions, which enable us to construct a designer-critic architecture. One LLM agent acts as the designer and proposes designs based on available information, while another agent acts as the critic to provide analysis and recommendation to encourage better designs. The agents interact sequentially through structured recommendations and a shared design history. Specifically, at each iteration, the designer proposes a batch of design candidates that follow three complementary strategies. 
\begin{itemize}
    \item refine: retain a promising mechanism and adjust a coherent group of parameters;
    \item redesign: make structural or parametric changes to address the main observed failure;
    \item alternative: explore a different arrangement or operating mechanism.

\end{itemize}

The design candidates are first passed to the programmatic validator and then used to generate geometric meshes for the task simulation in FEM. The simulation results are used to inform the next iteration. For the initial iteration, we leverage a reference design for the same family of tasks, such as a reference gripper for grasping tasks. 

Both agents receive the task specification, component information, and fitted module response models, but receive different feedback. The designer receives previous designs, task-level feedback from the FEM simulation, critic’s analysis and recommendation, and information from the programmatic validators identifying physical constraints that are violated. In addition, the critic receives this information together with detailed diagnostics and additional measurements from the privileged simulation. The critic first compares its previous predictions with measured outcomes, then proposes explanations for failures and provides recommendations on revisions corresponding to the refinement, redesign, and alternative strategies with confidence levels. The advice identifies the affected parameters, proposed changes, expected mechanism, predicted outcome, and relevant constraint margins. The designer then determines whether to follow these recommendations based on confidence levels. 

\section{Experiments
}
In this paper, we focus on bellow-shaped pneumatic soft robots as a representative example of complex soft robots that support parametric representation and manufacturing with 3D printing. Through the experiments, we aim to evaluate the performance and efficiency of the agentic task-driven design framework, the contribution of critic guidance and module response models, and the real-world performance of the agent-designed soft robots. 

\subsection{Parametric component families}

To evaluate the proposed framework, we construct a library containing two pneumatic component families: single-chamber bending modules with two different strain limiters, and three-chamber continuum actuators. The designer selects a component family and configures its geometry and spatial arrangement. Single-chamber modules with different geometric configurations can be combined into a monolithic finger, while components from either family can be selected as fingers mounted on a shared gripper palm. The bellows geometry is described by the connecting-tube outer radius \(r_t\), pneumatic bore radius \(r_b\), disc outer radius \(R_d\), nominal wall thickness \(t\), disc height \(h_d\), axial pitch \(\ell\), and convolution count \(n\). Here, \(\ell\) denotes the distance between corresponding points on adjacent discs. The bore radius and inter-disc gap are derived, \(r_b=r_t-t\) and \(g=\ell-h_d\).

\subsubsection{Single-chamber bending modules} This family consists of disc-shaped bellows connected through a central pneumatic chamber passage. A strain limiter restricts extension on one side, producing bending under pressurisation. The designer selects either a flat-back strain limiter or a fan-shaped filled sector with angular extent \(\psi\).

For the task-driven design, up to three modules can be joined as one finger, where the parameters can vary between different modules. The relative rotation \(\Delta\varphi\) between successive modules changes the orientation of their local bending directions. In this case, modules are joined into a monolithic structure with a continuous chamber, without intermediate plates. To preserve compatibility, tube dimensions and wall thickness are shared across the modules, and the complete finger is driven by one pressure input \(p\).

\subsubsection{Three-chamber continuum actuators} This family comprises three longitudinal chambers formed from pancake-shaped bellows around a solid triangular soft spine. Each chamber axis is offset from the component axis by a radial distance \(\rho\). The chambers retain a fixed angular separation of \(120^\circ\), while the designer can vary the overall chamber orientation \(\varphi_0\). Cavity dimensions are determined through the family-specific geometric parameterisation. For shape-matching tasks, the designer also selects the chamber-pressure vector \(\mathbf p=[p_1,p_2,p_3]^\mathsf T\), allowing bending direction to change through differential pressurisation.

For grasping, the designer can select either the single-chamber or the three-chamber module as fingers for a gripper, and mount them on a parametric palm with \(N_f\in\{2,3,4\}\) finger positions in radial or opposed arrangements. The designer selects the mounting radius \(R_a\), inclination \(\alpha\), finger configurations, and optional fingertip geometry. 

\subsection{Implementation and simulation settings}

In this paper, we leverage Claude Opus 5.5 with extra high thinking effort as the LLM model. To leverage the language capabilities of the agentic models, design proposals from the designer agent are expressed as typed JSON programs, which are used to generate CAD models with CadQuery, meshed with Gmsh, and evaluated in SOFA simulation with the SoftRobots plugin~\cite{geuzaine2009gmsh,faure2012sofa,coevoet2017software}. For the simulation of the soft robot body, we use corotational linear FEM, and a simulation time step of 0.02 s. The nominal mesh size is \(h=\min(3.5mm,2t)\), where \(t\) is wall thickness. The models use effective Young’s moduli of 1180 kPa for single-chamber components and 1250 kPa for three-chamber components, with Poisson’s ratio 0.44. To ensure printability, constraints on the wall-thickness, bore radius, finger length, robot body volume, and the number of finite elements are also imposed. Chamber actuation pressure is limited to 15 kPa.

\subsection{Evaluation}

We evaluate the agentic design framework using two downstream tasks that are typically used in prior works, including grasping and shape-matching, in SOFA simulation. The designs are subsequently demonstrated in the real world. 

\subsubsection{Grasping} For grasping, we use nine rigid primitive shape objects, including two egg-shaped objects weighing 150 and 190 g, 50 and 70 mm cubes, spheres, and pyramids, and a 50 mm diameter-by-height cylinder. These objects are selected based on the fact that the reference gripper used by the designer fails to grasp them successfully. Additionally, we leverage an additional, more challenging set of nine YCB objects that are typically seen in the real world, including tennis ball, tuna fish can, large marker, sponge, mug, flat screwdriver, scissors, strawberry, and foam brick~\cite{calli2017yalecmuberkeley}. The designer is tasked to design a gripper that can grasp the objects securely. 

The gripper design proposals are first validated and instantiated in SOFA simulation, where they are then evaluated in grasping experiments. Each grasp trial comprises the sequence of descent, closure, lift, shaking with 10 mm amplitude at 2 Hz, and settling. Each proposal is scored over three trials with 5\% perturbations to placement, friction, and pressure. The grasp score is computed as
\[ J_{\mathrm{grasp}} = \frac{1}{2}\operatorname{clip}_{[0,1]} \left(\frac{\Delta z_{\mathrm{lift}}}{40\ \mathrm{mm}}\right) + \frac{1}{2}\operatorname{clip}_{[0,1]} \left(\frac{\Delta z_{\mathrm{final}}}{40\ \mathrm{mm}}\right), \] where the displacements are relative to the initial object position. Geometrically blocked proposals are given zero scores without running the grasp simulation. The simulation results are then fed back to the designer and the critic agents for iterative improvements. The final best-performing design is then evaluated over ten separate additional noisy trials, and the reported success rate is the fraction of successful trials.

\subsubsection{Shape-matching} For shape-matching experiments, we leverage planar shapes such as s-bend shape and zigzag, and 3D shapes such as twist and multi-directional bending. Each design proposal is pressurised in free air, and its deformed centreline is compared with a $100\,\mathrm{mm}$ target curve. The centreline is taken as the cross-section centroid of the finger at stations every $10\,\mathrm{mm}$ along its rest length. Each station is compared with the target point at the same fraction of arc length, so a design must also match the target length. For $K$ targets, the objective is
\begin{equation}
    J_{\mathrm{shape}}
    = \frac{1}{K} \sum_{k=1}^{K}
    \max\left(0,\, 1 - \frac{\bar{e}_k}{40\,\mathrm{mm}}\right),
    \label{eq:shape_objective}
\end{equation}
where $\bar{e}_k$ is the mean distance between the deformed and target centrelines over the stations for target $k$.

For both the grasping tasks and the shape-matching tasks, we leverage three proposals for each iteration, and run for six iterations in total. This results in 19 total evaluated proposals for each task, including the initial reference design. Proposals that are blocked also count in these evaluations. The designer also selects the actuation pressure for each design, which is used as the target in a ramp controller to pressurise the robot. 

\subsection{Baselines and ablations}

For evaluation of the performance of the proposed agentic design framework, we compare with the following baselines in both the grasping and shape-matching tasks:

\begin{itemize}
    \item Random search: Independently samples candidate designs around the reference configurations using uniform samplers.
    \item Genetic algorithm (GA): Evolves design programs through mutation and crossover within an elitist search loop. Two elite designs are retained, and three proposals are generated per round, with selection based on simulation scores. Mutations to a finger are also applied to identical fingers in the assembly~\cite{holland1992adaptation}.
    \item Covariance Matrix Adaptation Evolution Strategy (CMA-ES): Optimises task-specific, fixed-length parameter vectors, initialised from the reference design. The population size is three, and the initial standard deviation is 0.3 of each parameter’s range~\cite{hansen2001completely}.
\end{itemize}

Additionally, to investigate the utility of the designer-critic framework and the module response model, we additionally include the following ablation experiments:

\begin{itemize}
    \item No critic guidance: Removes the critic agent. The designer revises candidates using validation results, simulation scores, and the task-level simulation results, without critic-generated analysis or recommendations.
    \item No module response models: Removes the fitted module response models, while keeping the designer-critic architecture. 
\end{itemize}

All methods use the same programmatic validator, simulation scores, and initial reference designs, with the same simulation evaluation budgets within each task.

\section{Results and Discussion}

This section discusses the results of the agentic framework for pneumatic soft robot design, and the performances of the designs in grasping and shape-matching tasks compared to baselines. Besides task successes, we also evaluate the efficiency of the design framework in obtaining working designs with the minimum number of simulation evaluations, which directly relate to the computational load required for running the SOFA simulations. For both the grasping and the shape-matching experiments, we leverage three seeds and compute the standard error of the mean.

\begin{figure} [h!]
	\centering
    \includegraphics[width=\columnwidth]{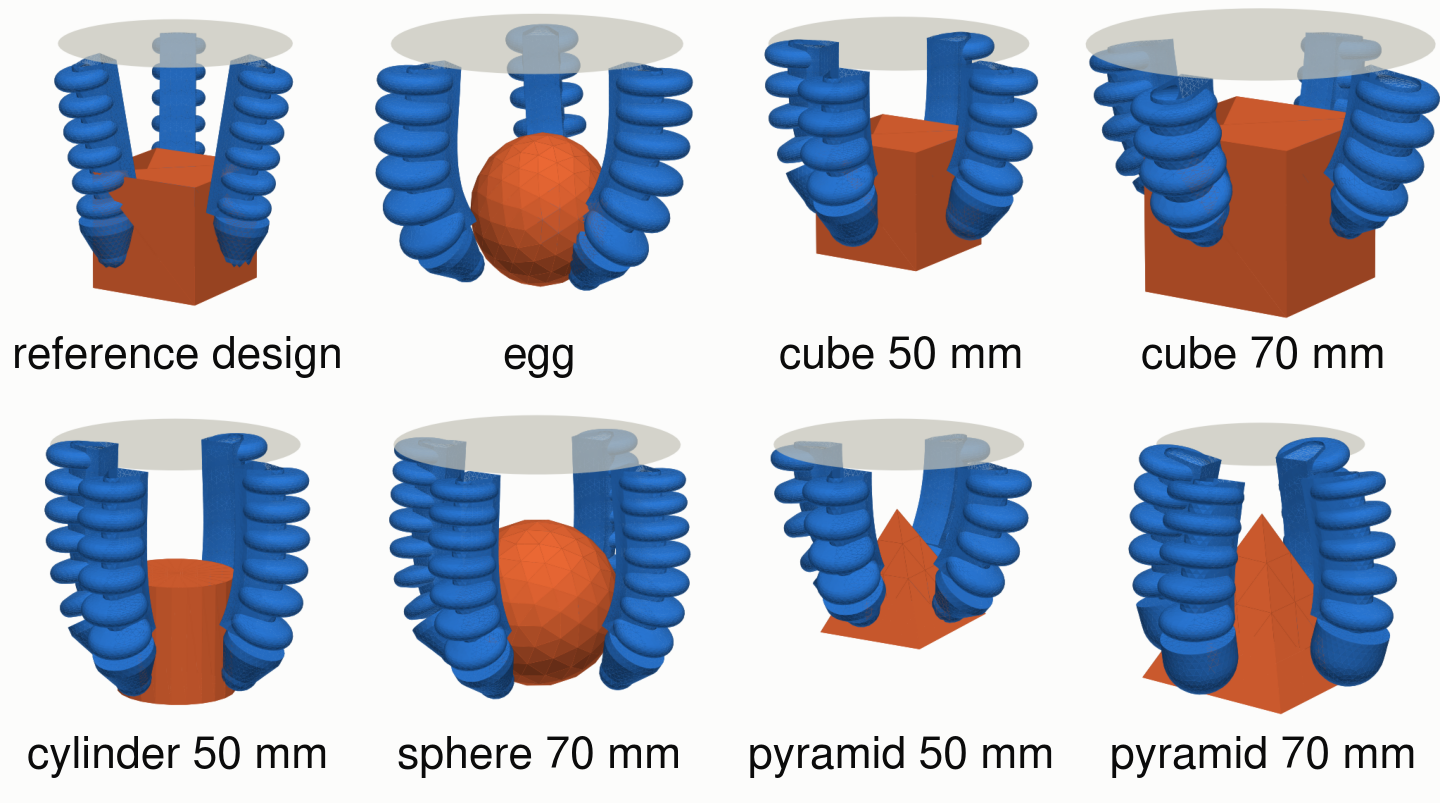}
	\caption{Illustration of agent-designed grippers in simulation.}
	\label{fig:gripper}
\end{figure} 

\subsection{Grasping}

The grasping performances are evaluated in the simulation using ten additional noisy trials on top of the challenging grasping procedures that include shaking. As can be seen in~\cref{tab:grasping_results}, in both primitive shape set and YCB object sets, the agentic design framework is able to significantly outperform the baselines in grasping performance, given the same evaluation budget in SOFA. One possible explanation for GA’s low performance is limited score variation among unsuccessful designs near the reference configuration.

Examples of the agent-designed grippers and the corresponding objects can be seen in \cref{fig:gripper}, which demonstrate that the agent is able to generate varied designs according to the requirement of the task. It should also be noted that these designs all share an initial reference design that fails on the task and require task-specific optimisations to be able to grasp the target objects. By incorporating prior physics knowledge from the module response models and their pre-training, as well as the ability to interpret simulation results and come up with dedicated solutions to grasping failures without manual involvement, the agentic framework demonstrates superior performance and improved efficiency.

\begin{table}[ht]
    \centering
    \small
    \setlength{\tabcolsep}{3pt}
    \caption{Mean simulated grasp hold rate $\pm$ SEM.}
    \label{tab:grasping_results}
    \begin{tabular}{lcccc}
        \toprule
        Object set & Agent & Random & GA & CMA-ES \\
        \midrule
        Prim $\uparrow$
        & $\mathbf{0.86 \pm 0.10}$
        & $0.15 \pm 0.08$
        & $0.04 \pm 0.04$
        & $0.31 \pm 0.10$ \\
        YCB $\uparrow$
        & $\mathbf{0.84 \pm 0.04}$
        & $0.17 \pm 0.07$
        & $0.07 \pm 0.05$
        & $0.38 \pm 0.08$ \\
        \bottomrule
    \end{tabular}
\end{table}

Furthermore, to further contextualise the performance of the agentic framework against the baselines, we perform an additional experiment by running the computational baseline methods with more than six times the FEM evaluation budget by increasing the population size and the number of iterations. On the primitive shape objects, CMA-ES achieves a hold rate of 0.88$\pm$0.12, similar to the agentic framework, and GA achieves 0.17$\pm$0.17. On YCB objects, CMA-ES only achieves 0.62$\pm$0.18 and GA achieves a hold rate similar to that obtained on primitive shapes, both remaining significantly below the agentic framework. This demonstrates that PneuGen is able to find high-performing designs much more efficiently than these baselines, and is also more capable of handling complex shapes.

\subsection{Shape-matching}

\begin{figure} [h]
	\centering
    \includegraphics[width=\columnwidth]{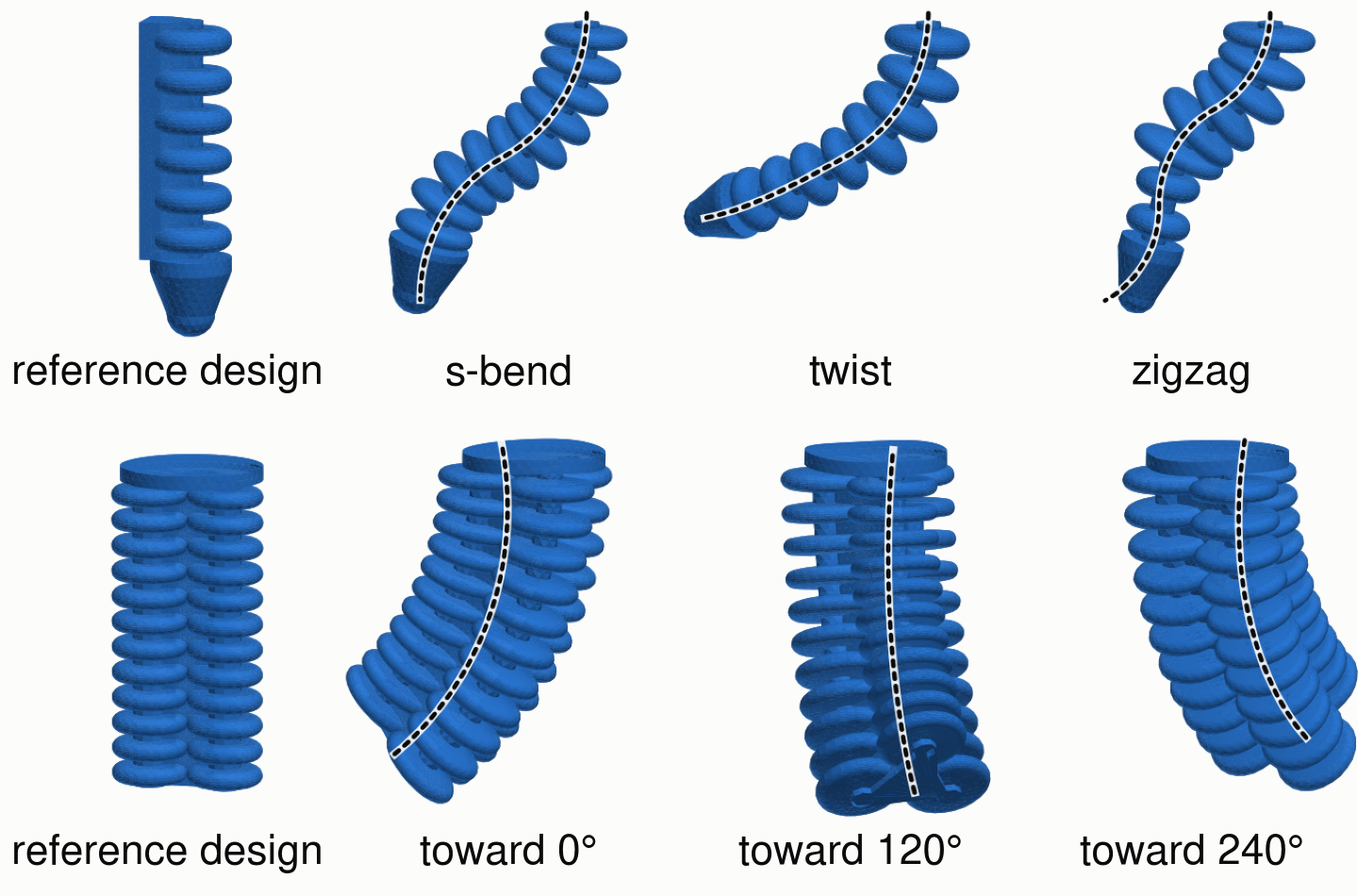}
	\caption{Illustration of shape-matching experiments in simulation. Target centrelines are marked with a dotted line.}
	\label{fig:shape-matching}
\end{figure} 

In shape-matching tasks, the agentic framework is also able to outperform the baseline methods and design pneumatic soft robots that are better able to produce the desired shapes. As seen in \cref{tab:shape_matching_results}, across the different desired planar and 3D shapes, the framework’s designs achieve a lower mean centreline error, demonstrating effective use of module behaviour information. It should also be noted that in the shape-matching task, the task objectives provide an arguably stronger signal than the grasp score, especially for the computational optimisation baseline methods; yet the agentic framework is still able to outperform them given the same evaluation budget. Visual demonstrations of the shape-matching experiments are also shown in \cref{fig:shape-matching}. In tasks such as the s-bend, the generated designs used changes in local bending orientation between successive modules to reproduce opposing curvatures under a shared pressure input. Furthermore, as some of the shape-matching tasks require the three-chamber actuator instead of the single-chamber module, this demonstrates that the framework is able to connect the component behaviour with the requirement of the task and select the appropriate component family for iterative optimisation. 

\begin{table}[h]
    \centering
    \small
    \setlength{\tabcolsep}{2pt}
    \caption{Mean shape-matching score $\pm$ SEM}
    \label{tab:shape_matching_results}
    \begin{tabular}{@{}lcccc@{}}
        \toprule
        Task & Agent & Random & GA & CMA-ES \\
        \midrule
        Planar\,$\uparrow$
        & $\mathbf{0.95 \pm 0.02}$
        & $0.82 \pm 0.03$
        & $0.78 \pm 0.01$
        & $0.83 \pm 0.02$ \\
        3D\,$\uparrow$
        & $\mathbf{0.94 \pm 0.03}$
        & $0.72 \pm 0.02$
        & $0.76 \pm 0.01$
        & $0.74 \pm 0.00$ \\
        \bottomrule
    \end{tabular}
\end{table}

\subsection{Ablations}
To investigate the utility of the different building blocks of the agentic framework, we evaluate two ablations: removing the critic agent and removing the fitted module response models. As seen in \cref{tab:ablation}, using the critic agent is able to improve the performance of the soft robot design in both the grasping tasks and the shape-matching tasks. In the grasping task, the critic improvement is also slightly more significant for the more challenging YCB objects, pointing to a potential for improving designs for more complex objects. Additionally, using the critic agent enables the framework to achieve task successes with fewer FEM evaluations than using the designer agent alone, reducing the average number of required evaluations by two and six for grasping and shape-matching tasks respectively. Through stronger diagnostics and analysis, the critic agent helps avoid design failures, and reduces the computational load for running expensive FEM simulations. 

By examining the designer’s recorded design rationales, we identify that the critic agent is typically able to suggest better designs. The critic is also tasked to focus on analysing failures in the design proposals, thus potentially providing more insights into the reasons behind task failures for the designer to consider. Additionally, the incorporation of the confidence label also prevents forcing the designer to follow the critic’s recommendations. 

\begin{table}[t]
    \centering
    \small
    \setlength{\tabcolsep}{2pt}
    \caption{Ablation: mean grasp hold rate and shape-matching score $\pm$ SEM.}
    \label{tab:ablation}
    \begin{tabular}{@{}lccc@{}}
        \toprule
        Task & Agent & w/o critic & w/o model \\
        \midrule
        Grasping (primitives)\,$\uparrow$
        & $\mathbf{0.86 \pm 0.10}$
        & $0.79 \pm 0.11$
        & $0.81 \pm 0.13$ \\
        Grasping (YCB)\,$\uparrow$
        & $\mathbf{0.84 \pm 0.04}$
        & $0.74 \pm 0.07$
        & $0.76 \pm 0.11$ \\
        Shape-match (planar)\,$\uparrow$
        & $\mathbf{0.95 \pm 0.02}$
        & $0.88 \pm 0.01$
        & $0.92 \pm 0.03$ \\
        Shape-match (3D)\,$\uparrow$
        & $\mathbf{0.94 \pm 0.03}$
        & $0.90 \pm 0.00$
        & $0.92 \pm 0.01$ \\
        \bottomrule
    \end{tabular}
\end{table}

From results of the ablation without the response model in \cref{tab:ablation}, it can be seen that removing the module response model reduces the performance of the agentic design framework. One possible explanation is that, without such models, the agents need to first perform explorations to understand the behaviour of the modules when actuated. LLM agents offer an intuitive way to incorporate such reusable models of module behaviour, which is typically unavailable for baselines without extensive manual efforts in incorporating the models into the computational search. However, the improvement is less significant than the critic, potentially because the module behaviour is relatively simple and can be sufficiently explored within the evaluation budget. Including the response model also reduces the number of FEM evaluations required to achieve task success by two and four for grasping and shape-matching respectively, thus making the design process more efficient. In this way, the PneuGen framework offers a pathway for efficient design of soft robots leveraging libraries of components with known behaviours. 

\subsection{Real world demonstration}

\begin{figure}[!tbp]
  \centering
  \subfloat[]{\includegraphics[width=0.0985\textwidth]{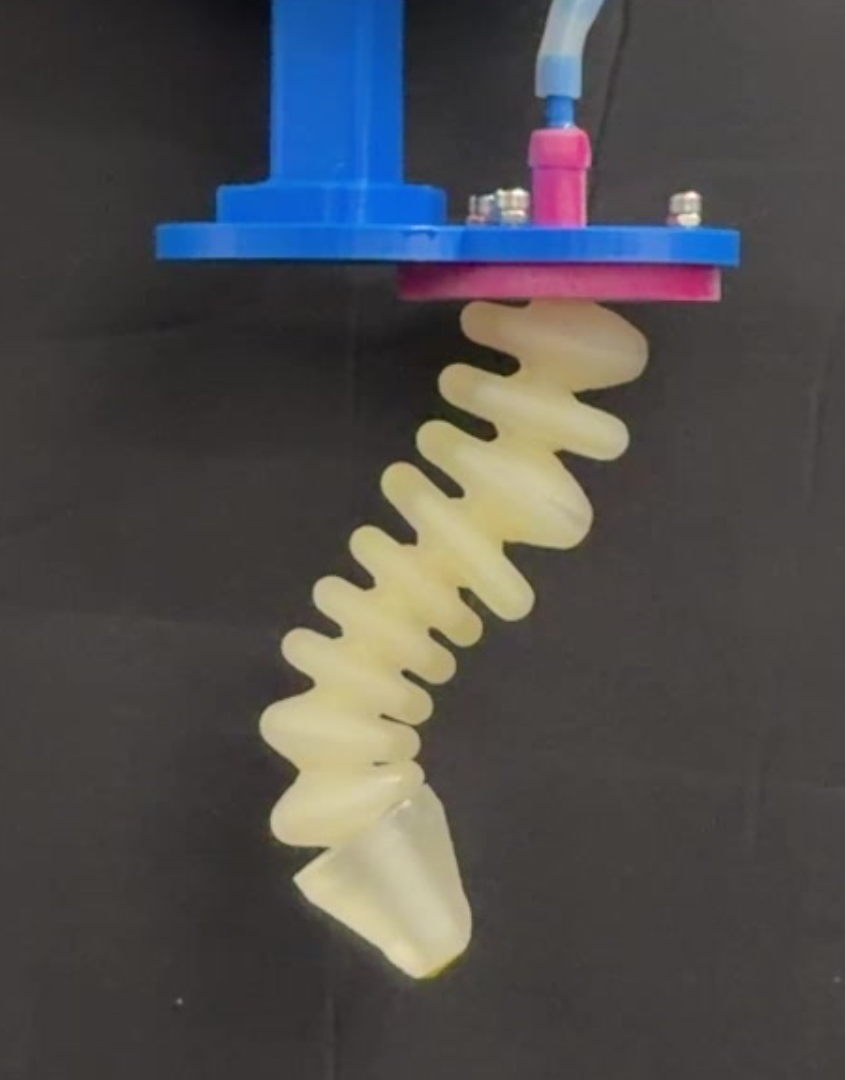}\label{fig:real:shape}}
  \hfill
  \subfloat[]{\includegraphics[width=0.385\textwidth]{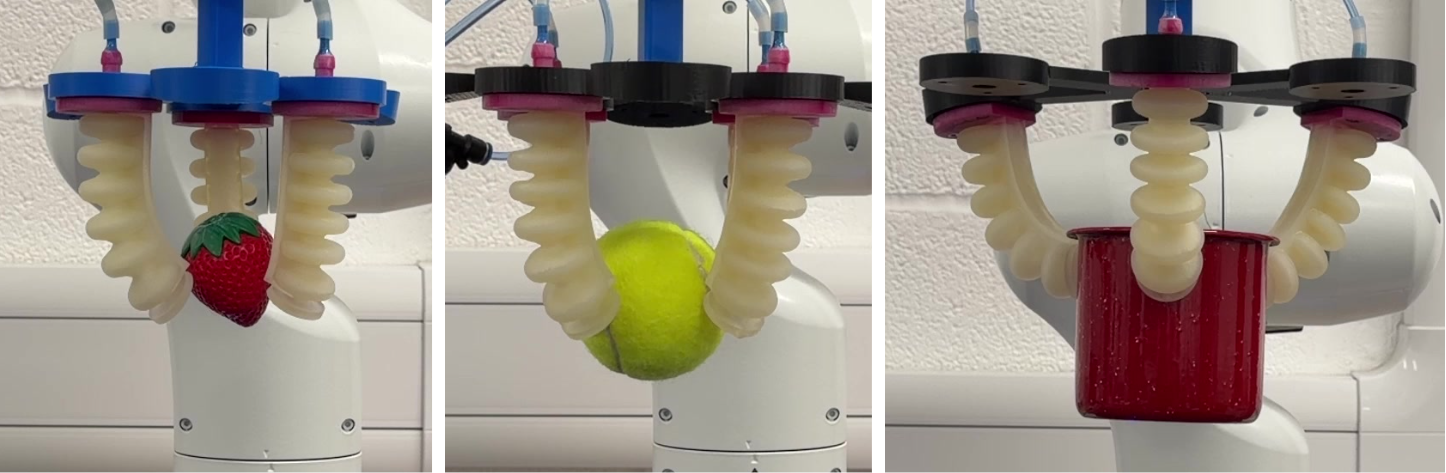}\label{fig:real:grasp}}
  \caption{Real-world demonstration of agent-designed robots: a) shape-matching for s-bend; b) YCB object grasping.}
  \label{fig:real}
\end{figure}

To validate the PneuGen framework in the real world, we 3D printed the agent-designed soft robots for the shape-matching and object grasping tasks. As can be seen in \cref{fig:real}, the printed soft pneumatic robots are able to match the desired shape and grasp the target objects successfully. In real-world grasping evaluations, the agent-designed grippers are also able to successfully  execute target behaviours and grasp the real world YCB objects securely without failure. These results demonstrate the validity of the simulation experiment evaluations, and also illustrate the utility of the PneuGen framework in efficient task-driven design of pneumatic soft robots for real world applications. 
\section{Conclusion}
We presented PneuGen, a task-driven co-design framework for pneumatic soft robots that combines a designer LLM agent with a critic agent for iterative design optimisation. The framework uses a parametric component library, module response models, and automatic CAD-to-FEM generation to support the joint selection of component family, spatial arrangements, geometric parameters, and actuation pressures. Across grasping and shape-matching tasks in FEM simulation, PneuGen outperforms all baselines under matched evaluation budgets. Ablations also show that critic guidance and module response models improved performance and reduced the number of FEM evaluations needed to achieve task success. We further validate the PneuGen framework by 3D printing agent designs, which demonstrate successful grasping and shape-matching in the real world. Future work should extend the component library to include additional actuation mechanisms, and also evaluate the use of different LLMs as agents.

\addtolength{\textheight}{-12cm}   


\bibliographystyle{IEEEtran}
\bibliography{ieee-control,references}

\end{document}